\documentclass[10pt,twocolumn,letterpaper]{article}

\usepackage{cvpr}              

\usepackage{graphicx}
\usepackage{amsmath}
\usepackage{amssymb}
\usepackage{booktabs}

\usepackage{url}
\usepackage[ruled,vlined,linesnumbered]{algorithm2e}
\usepackage{bbm}
\usepackage{multirow}
\usepackage{booktabs}
\usepackage{algpseudocode}
\usepackage{subcaption}

\usepackage[pagebackref,breaklinks,colorlinks]{hyperref}

\usepackage[capitalize]{cleveref}
\crefname{section}{Sec.}{Secs.}
\Crefname{section}{Section}{Sections}
\Crefname{table}{Table}{Tables}
\crefname{table}{Tab.}{Tabs.}

\def\cvprPaperID{*****} 
\def\confName{CVPR}
\def\confYear{2023}

\newtheorem{definition}{Definition}

\newcommand{\blue}[1]{\textcolor{blue}{#1}}

\newtheorem{lemma}{Lemma}

\usepackage{xcolor}
\definecolor{lightblue}{RGB}{73, 116, 150}
\definecolor{codekey}{RGB}{255,140,0}
\definecolor{deepblue}{RGB}{25, 60, 100}
\newcommand{\lightblue}[1]{\textcolor{lightblue}{#1}}

\newcommand{\deepblue}[1]{\textcolor{deepblue}{#1}}

\title{ImbaGCD: Imbalanced Generalized Category Discovery}

\author{
Ziyun Li\\
Hasso Plattner Institute\\
University of Potsdam\\
{\tt\small ziyun.li@hpi.de}
\and
Ben Dai\\
Chinese University of HongKong\\
{\tt\small bendai@cuhk.edu.hk} 
\and
Furkan Simsek\\
Hasso Plattner Institute\\
University of Potsdam\\
{\tt\small 	furkan.simsek@hpi.de}
\and
Christoph Meinel\\
Hasso Plattner Institute\\
University of Potsdam\\
{\tt\small 	christoph.meinel@hpi.de}
\and
Haojin Yang\\
Hasso Plattner Institute\\
University of Potsdam\\
{\tt\small 	haojin.yang@hpi.de}
}

\begin{document}
\maketitle

\begin{abstract}
   {Generalized class discovery (GCD) aims to infer known and unknown categories in an unlabeled dataset leveraging prior knowledge of a labeled set comprising known classes. 
   Existing research implicitly/explicitly assumes that the frequency of occurrence for each category, whether known or unknown, is approximately the same in the unlabeled data. 
   However, in nature, we are more likely to encounter known/common classes than unknown/uncommon ones, according to the long-tailed property of visual classes. 
   Therefore, we present a challenging and practical problem, Imbalanced Generalized Category Discovery (ImbaGCD), where the distribution of unlabeled data is imbalanced, with known classes being more frequent than unknown ones. 
   To address these issues, we propose ImbaGCD, A novel optimal transport-based expectation maximization framework that accomplishes generalized category discovery by aligning the marginal class prior distribution.
   ImbaGCD also incorporates a systematic mechanism for estimating the imbalanced class prior distribution under the GCD setup.
   Our comprehensive experiments reveal that ImbaGCD surpasses previous state-of-the-art GCD methods by achieving an improvement of approximately \textbf{2 - 4\%} on CIFAR-100 and \textbf{15 - 19\%} on ImageNet-100, indicating its superior effectiveness in solving the Imbalanced GCD problem.
   }

\end{abstract}

\section{Introduction}




\begin{figure}[t]
   \vspace{0.7cm}
   \centering
   \includegraphics[width=1\linewidth]{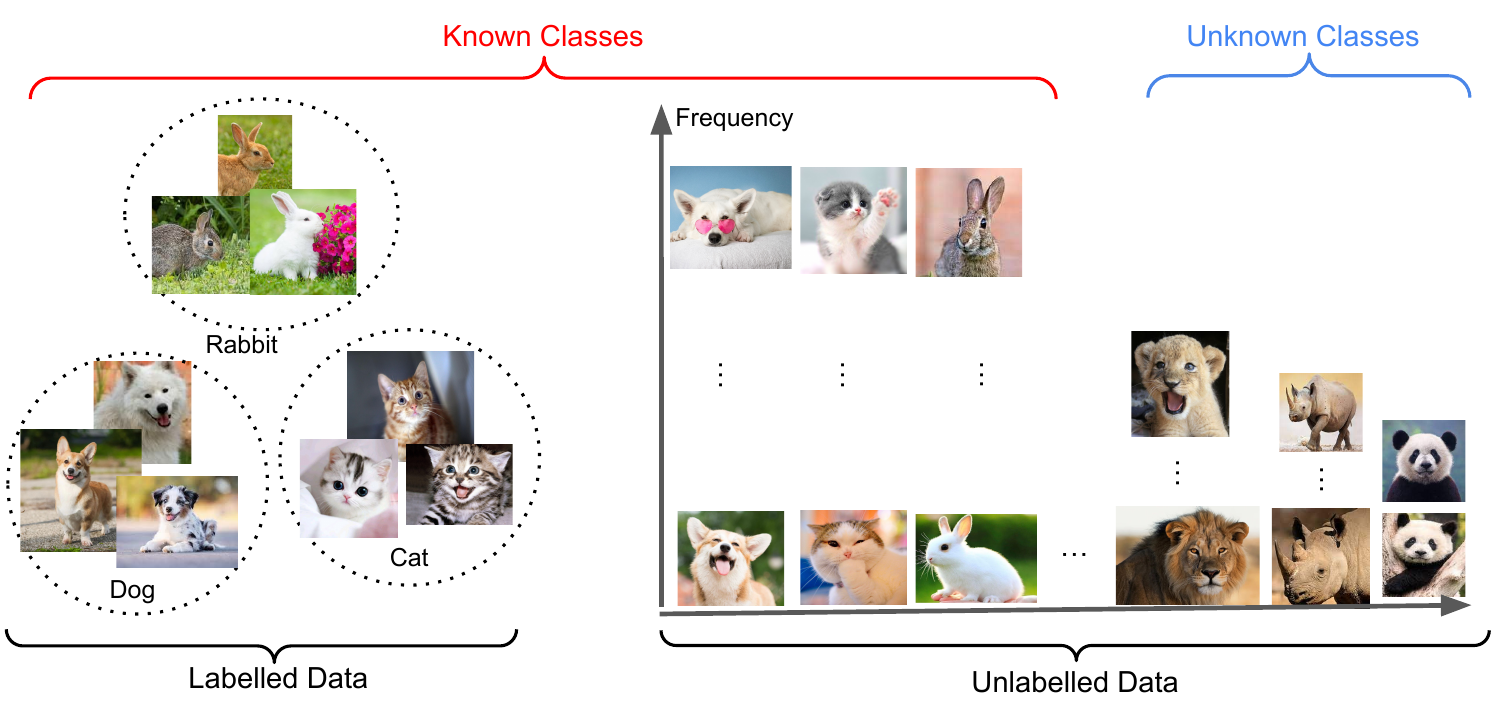}
   \caption{Illustration of ImbaGCD. 
   ImbaGCD attempts to identify known categories and discover new classes within a vast amount of unlabeled data from the real world.
   In the real world, unlabeled data includes both known class from the labeled set and unknown class, where known classes (e.g., cat, dog, rabbit) dominate the well-represented ``head," while ``unknown" classes (e.g., lion, rhino, panda) primarily reside in the underrepresented ``tail" of the distribution.}
   \vspace{-3mm} 
   \label{fig:ImbaGCD}
\end{figure}

Existing machine learning models can attain excellent performance  when trained on large-scale datasets with human annotations. 
However, the success of these models is strongly dependent on the fact that they are only required to recognize images from the same set of classes with extensive human annotations on which they are train.
This constraint restricts the applicability of these models in the real world where unannotated data from unseen categories may be encountered.
To overcome this limitation, researchers have developed related research topics, such as semi-supervised learning \cite{chapelle2009semi, hady2013semi} that utilizes both labeled and unlabeled data to train a robust model, few-shot learning \cite{snell2017prototypical} that aims to generalize to new classes with limited annotated samples, open-set recognition \cite{scheirer2012toward} that identifies whether an unlabeled image belongs to one of the known classes, and novel category discovery (NCD) \cite{han2019learning,han2021autonovel,fini2021unified,li2022closer} that partitions unannotated data from unknown categories by transferring knowledge from known ones.
NCD initially assumed all unannotated images were from unknown categories, which is not realistic. 
GCD \cite{vaze2022generalized} is introduced to consider unannotated images from both known and unknown categories, better reflecting real-world scenarios.

The Generalized Category Discovery (GCD) problem presents challenges due to the agnostic nature of real-world unlabeled data. 
Two primary concerns arise: 
(i) \textit{limited knowledge of new categories' appearances}, which complicates selecting labeled datasets with high semantic similarity. 
As indicated in \cite{li2023sup, li2022closer,chi2021meta}, GCD may become ill-defined if known and novel classes do not share high-level semantic features. 
Furthermore, \cite{li2023sup, li2022closer} highlights that utilizing supervised knowledge from labeled sets can result in suboptimal performance in low semantic similarity situations.
(ii) \textit{Lack of information about the occurrence frequency} for each category, whether known or unknown, which constitutes the central focus of our research.
However, existing research \cite{vaze2022generalized, sunopencon, cao2022open, rizve2022openldn} often overlooks the issue of category occurrence frequency and tends to use experimental setups based on \cite{vaze2022generalized}, where the unknown class occurs twice as often as the known class. This setting, however, misrepresents real-world scenarios. In line with the long-tailed property of visual classes, we are more likely to encounter known classes in natural environments. As shown in Figure \ref{fig:ImbaGCD}, known classes (e.g, cat, dog and rabbit)  with their ease of label acquisition, dominate the well-represented ``head'' of the distribution, while ``unknown'' classes (e.g., lion, rhino and panda) are harder to obtain, primarily residing in the underrepresented ``tail''.

Therefore, we present \underline{Imba}lanced \underline{G}eneralized \underline{C}ategory \underline{D}iscovery (ImbaGCD), a realistic problem where the distribution of unlabeled data is imbalanced, with known classes more frequent than unknown ones. Our work focuses on two primary challenges:
(i) \textit{Estimating indeterminate class prior.} Unlike conventional long-tailed learning (LTL) scenarios \cite{buda2018systematic,liu2019large,cao2019learning}, most existing methods \cite{chawla2002smote,he2009learning,huang2016learning,cui2019class,kang2020decoupling} rely on supervised learning and labeled data for handling class imbalance. 
This renders the estimation of the marginal class prior distribution particularly challenging, as it cannot be achieved by simply counting training samples based on class labels.
In this work, we address this by an iterative class prior estimation technique, which serves as a robust approximation of the true prior (Section \ref{subsec:prior_estimation}). 
(ii) \textit{Mitigating bias towards head classes,} which is a key challenge in imbalanced data settings \cite{khan2018cost}, and is also a primary concern in ImbaGCD. 
Specifically, tail/unknown samples are often misclassified as head/known classes, as models tend to focus on learning patterns from prevalent classes, leading to poorer performance on minority classes.
Our approach involves applying constraints to pseudo-labels to align their distribution with the class prior distribution, promoting the identification of unlabeled samples as unknown classes. By formulating this as an optimal transport problem \cite{villani2008optimal}, we efficiently solve the constrained optimization objective using the Sinkhorn-Knopp algorithm \cite{cuturi2013sinkhorn}.

We carriy out a comprehensive evaluation of ImbaGCD on benchmark datasets, outperforming the state-of-the-art by margins of approximately {2-4\%} and {15-19\%} on CIFAR-100 and ImageNet-100, respectively, across multiple imbalanced settings. In addition, ImbaGCD exhibits competitive performance in both balanced and original GCD settings, underscoring its versatility and effectiveness in a variety of situations.
Our contributions are summarized as follows:
\begin{itemize}
    \item We present a challenging and practical problem, Imbalanced Generalized Category Discovery (ImbaGCD), where the distribution of unlabeled data is imbalanced, with known classes being more frequent than unknown ones.
    \item We propose ImbaGCD, a novel optimal transport-based expectation maximization framework that enables the discovery of generalized classes by matching the marginal class prior distribution. ImbaGCD also incorporates a systematic mechanism for estimating the imbalance class prior distribution under the GCD setup.
    \item Our extensive experiments demonstrate that ImbaGCD outperforms previous state-of-the-art GCD methods on standardized benchmarks, indicating its superior effectiveness in solving the ImbaGCD problem.
\end{itemize}
\section{PRELIMINARIES}
In this section, we first introduce the GCD setup, and then briefly review the optimal transport.
\subsection{Problem Setup}
We denote $(\mathbf{X}_l, Y_l)$ and $(\mathbf{X}_u, Y_u)$ as random samples under the \emph{labeled/unlabeled probability measures} $\mathbb{P}_{\mathbf{X}, Y}$ and $\mathbb{Q}_{\mathbf{X}, Y}$, respectively.
$\mathbf{X}_l \in \mathcal{X}_l \subset \mathbb{R}^d$ and $\mathbf{X}_u \in \mathcal{X}_u \subset \mathbb{R}^d$ are the labeled/unlabeled feature vectors, $Y_l \in \mathcal{Y}_l$ and $Y_u \in \mathcal{Y}_u$ are the true labels of labeled/unlabeled data, where $\mathcal{Y}_l$ and $\mathcal{Y}_u$ are the label sets under the labeled and unlabeled probability measures $\mathbb{P}_{\mathbf{X}, Y}$ and $\mathbb{Q}_{\mathbf{X}, Y}$, respectively.
\begin{definition}[Generalized Category Discovery]
Let $\mathbb{P}_{\mathbf{X}_l,Y_l}$ be a labeled probability measure on $\mathcal{X}_l \times \mathcal{Y}_l$, and $\mathbb{Q}_{\mathbf{X}_u,Y_u}$ be an unlabeled probability measure on $\mathcal{X}_u \times \mathcal{Y}_u$, with $\mathcal{Y}_l\subset\mathcal{Y}_u$. ($\mathcal{Y}_u$ comprises both known classes $\mathcal{Y}_l$ and unknown classes $\mathcal{Y}_n=\mathcal{Y}_u\backslash\mathcal{Y}_l$.).
Given a labeled dataset $\mathcal{L}_n$ sampled from $\mathbb{P}_{\mathbf{X}_l, Y_l}$ and an unlabeled dataset $\mathcal{U}_m$ sampled from $\mathbb{Q}_{\mathbf{X}_u}$, GCD aims to predict the label $Y_u$ for each unlabeled instance $X_u$, which may belong to either known or unknown classes.
\end{definition}  
\vspace{-1.8mm}
Specifically, under our proposed imbalanced GCD, as discussed before, the class distributions in the unlabeled set are skewed with  $\mathbb{P}_{Y_{{l}}}(i) > \mathbb{P}_{Y_{{n}}}(j)$ for all $i \in \mathcal{Y}_l$ and $j \in \mathcal{Y}_n$. This problem formulation acknowledges the class distribution differences between known and unknown categories in unlabeled sets, instead of assuming $\mathbb{P}_{Y_{{l}}}(i) \approx \mathbb{P}_{Y_{{n}}}(j)$  for all $i \in \mathcal{Y}_l$ and $j \in \mathcal{Y}_n$.

\subsection{Reminders on Optimal Transport}
\label{sec:ot}

Optimal transport (OT) is a method to quantify the cost of converting one probability measure to another, which offers a unique perspective to understand imbalanced problems. Given two random variables $\mathbf{X}$ and $Y$, their corresponding probability measures are denoted as $\mathbf{r}$ and $\mathbf{w}$. 
Furthermore, the cost function $\mathcal{C}(\mathbf{X}, Y): \mathbf{X} \times Y \rightarrow \mathbb{R}_{+}$ represents the expense of transferring $\mathbf{X}$ to $Y$. Consequently, the OT distance between $\mathbf{X}$ and $Y$ can be defined as follows:
\begin{gather*}
   \text{OT} (\mathbf{r}, \mathbf{w})=\min _{\boldsymbol{\pi} \in \Pi(\mathbf{r}, \mathbf{w})} \int_{\mathbf{X} \times Y} \mathcal{C}(\boldsymbol{x}, {y}) \pi(\boldsymbol{x}, {y}) d \boldsymbol{x} d {y} \\
\Pi(\mathbf{r}, \mathbf{w}):=\left\{\int_{{Y}} \pi(\boldsymbol{x}, {y}) d{y}=\mathbf{r}(\boldsymbol{x}), \int_{X} \pi(\boldsymbol{x}, {y}) d \boldsymbol{x}=\mathbf{w}(y)\right\}
\end{gather*}
where $\pi(\mathbf{r}, \mathbf{w})$ is the joint probability measure with $\mathbf{r}$ and $\mathbf{w}$ \cite{villani2009optimal}. 
In empirical version, the OT distance can be expressed using discrete distributions and a cost matrix $\mathbf{M}$:
\begin{gather*}
   d_{M}(\mathbf{r}, \mathbf{w})=\min _{\mathbf{P} \in U(\mathbf{r}, \mathbf{w})}\langle\mathbf{P}, \mathbf{M}\rangle \\
U(\mathbf{r}, \mathbf{w}):=\left\{\mathbf{P} \in \mathbb{R}_{+}^{d \times d} \mid \mathbf{P} \mathbf{1}_{d}=\mathbf{r}, \mathbf{P}^{T} \mathbf{1}_{d}=\mathbf{w}\right\}
\end{gather*}
where $U(\mathbf{r}, \mathbf{w})$ represents the transport polytope of $\mathbf{r}$ and $\mathbf{w}$, namely the polyhedral set of $d \times d$ matrices with non-negative entries whose rows and columns sum to $\mathbf{r}$ and $\mathbf{w}$, respectively.
The primary objective of OT is to determine a transportation matrix $\mathbf{P}$ that minimizes the distance $d_{M}(\mathbf{r}, \mathbf{w})$.
Although OT serves as a distance measure between probability distributions under a specific cost matrix, solving the optimization problem with network simplex or interior point methods can be computationally demanding. To address this challenge, OT with entropy constraint has been introduced, which optimizes with a lower computational cost while maintaining sufficient smoothness \cite{cuturi2013sinkhorn}. By incorporating a Lagrangian multiplier for the entropy constraint, the new formulation is defined as:
\begin{gather*}
   d_{\mathbf{M}}^{\lambda}(\mathbf{r}, \mathbf{w})=\left\langle\mathbf{P}^{\lambda}, \mathbf{M}\right\rangle \\ 
   \quad \text {where} \quad \mathbf{P}^{\lambda}=\underset{\mathbf{P} \in U(\mathbf{r}, \mathbf{w})}{\arg \min }\langle\mathbf{P}, \mathbf{M}\rangle-\lambda h(\mathbf{P}),
\end{gather*}
$\lambda>0$, $h(\mathbf{P})=-\sum_{n=1}^{N} \sum_{k=1}^{K} \mathbf{P}_{nk} \log \mathbf{P}_{nk}$, and $d_{\mathbf{M}}^{\lambda}(\mathbf{r}, \mathbf{w})$ is also referred to as dual-Sinkhorn divergence. The matrix scaling algorithms can compute this divergence with reduced computational demand. A lemma ensures the convergence and uniqueness of the solution.

\begin{lemma}
   For $\lambda > 0$, the solution $\mathbf{P}^{\lambda}$ is unique and can be represented as $\mathbf{P}^{\lambda} = \operatorname{diag}(\mathbf{\alpha}) \mathbf{K} \operatorname{diag}(\mathbf{\beta})$, where $\mathbf{\alpha}$ and $\mathbf{\beta}$ are two non-negative vectors uniquely determined up to a multiplicative factor, and $\mathbf{K} = e^{-\mathbf{M}/ \lambda}$ is the element-wise exponential of $-\mathbf{M} / \lambda$.
   \end{lemma}
   The lemma above establishes the uniqueness of the solution $\mathbf{P}^{\lambda}$ \cite{sinkhorn1974diagonal}. Additionally, $\mathbf{P}^{\lambda}$ can be efficiently computed using Sinkhorn's fixed point iteration: $\mathbf{\alpha}, \mathbf{\beta} \leftarrow \mathbf{r} . / \mathbf{K} \mathbf{\beta}, \mathbf{w} . / \mathbf{K}^{\top} \mathbf{\alpha}$, where $. /$ denotes element-wise division.

\begin{algorithm}
   \SetAlgoLined
   \small
   \KwIn{Training dataset D, classifier $f$, uniform marginal $\mathbf{r}$, $\mathbf{w}$, and hyperparameters $\lambda_{proto}, \lambda_{sup}, \tau, \mu$}
   \SetKwFunction{algofunc}{ImbaGCD}
   \SetKwProg{myalg}{Algorithm}{:}{}
   \myalg{\algofunc{$\lambda_{proto}, \lambda_{sup}, \tau, \mu$}}{
       \For{epoch = $1,2,...,$} 
       {
         \For{step = $1,2,...,$} 
         {
        \lightblue{/ /E-step: estimate pesudo-label matrix $\mathbf{A}$:} \\
            Get classifier prediction $\mathbf{P}$ on a mini-batch of the unlabeled data of size $B_u$ \\
            Calculate $\mathbf{K}$ such that $k_{ij} = p_{ij}^\lambda$ \\ 
            \For{t=$1,...,T$}
            {
                \lightblue{/ / Sinkhorn’s fixed point iteration\\}       
                $\mathbf{\alpha} \leftarrow \mathbf{w} . /(\mathbf{K} \mathbf{\beta}), \quad \mathbf{\beta} \leftarrow \mathbf{r} . /\left(\mathbf{K}^{\top} \mathbf{\alpha}\right)$ 
            }
            $\mathbf{A}= B_u diag(\mathbf\alpha)\mathbf{K}diag(\mathbf\beta)$ \\
            \lightblue{/ / M-step: update model parameters:} \\
         $\theta^{k}=SGD(\mathcal{L_\text{overall}}, \theta^{k-1})$~~~~~~~~~~~~~\\
         \lightblue{/ / Overall loss function is provided Eq.\ref{eq:overall}}
         }
         $\mathbf{r} \leftarrow \mu \mathbf{r}+(1-\mu) \mathbf{z}$\lightblue{/ /update the class prior} \\
         $\mathbf{c}_{k} \leftarrow \mu \mathbf{c}_{k}+(1-\mu) \mathbf{v}_k$\lightblue{/ /update the prototype} \\

      }
   }
   \caption{Pseudo-code for ImbaGCD}
   \label{algo}
\end{algorithm}

\section{Method}
Our proposed method, ImbaGCD, is based on the Expectation-Maximization (EM) algorithm, which assigns pseudo-labels using the Sinkhorn algorithm in the E-step and updates the model with contrastive learning in the M-step. We iteratively update the class prior and prototype every epoch. The detailed process is outlined in Algorithm \ref{algo}.
\label{sec:sinkEM}
\subsection{Prototype Loss for Unlabeled Data} 
\label{sec:prototype_loss}
Our objective is to find the network parameters $\theta$ that maximizes the log-likelihood function of the observed $m$ unlabeled samples:
$\small
   \theta^{*}=\underset{\theta}{\arg \max } \sum_{i=1}^{m} \log p\left(\mathbf{x}^{(i)} \mid \theta\right).
$
We assume that the observed data $\left\{\mathbf{x}^{(i)}\right\}_{i=1}^{m}$ are related to latent variable $\mathbf{C}=\left\{\mathbf{c}_{k}\right\}_{k=1}^{|\mathcal{Y}_u|}$ which denotes the prototypes of the data, where $|\mathcal{Y}_u|$ denotes the total number of unknown classes ($\mathcal{Y}_l \subset \mathcal{Y}_u$). The joint distribution of $\mathbf{x}^{(i)}$ and $\mathbf{c}^{(i)}$ is given as follow:
%

\begin{equation}
    \small
       \begin{aligned}
          \theta^{*} = \underset{\theta}{\arg \max } \sum_{i=1}^{m} \log \sum_{c_{k} \in C} p\left(\mathbf{x}^{(i)}, \mathbf{c}_{k}^{(i)} \mid \theta\right)
       \end{aligned}   
    \end{equation}
It is hard to optimize this function directly, so we use a surrogate function to lower-bound it:
%
%
\begin{equation}
\label{eq:lowerbound}
\begin{aligned}
   &\sum_{i=1}^{m} \log \sum_{c_{k} \in C} p\left(x^{(i)}, c_{k}^{(i)} \mid \theta\right) \\
   =& \sum_{i=1}^{m} \log \sum_{c_{k} \in C} Q\left(c_{k}^{(i)}\right) \frac{p\left(x^{(i)}, c_{k}^{(i)} \mid
   \theta\right)}{Q\left(c_{k}^{(i)}\right)}  \\
   \geq& \sum_{i=1}^{m} \sum_{c_{k} \in C} Q\left(c_{k}^{(i)}\right) \log \frac{p\left(x^{(i)}, c_{k}^{(i)} \mid \theta\right)}{Q\left(c_{k}^{(i)}\right)}
\end{aligned}
\end{equation}

%
%
To achieve equality, we have
$
Q\left(\mathbf{c}_{k}^{(i)}\right)=p\left(\mathbf{c}_{k}^{(i)} \mid \mathbf{x}^{(i)}, \theta\right)
$.
%
%
By ignoring the constant, we aim to maximize:
\begin{equation}\label{eq:em2}
\small
   \sum_{i=1}^{m} \sum_{c_{k} \in C} Q\left(\mathbf{c}_{k}^{(i)}\right) \log p\left(\mathbf{x}^{(i)}, \mathbf{c}_{k}^{(i)} \mid \theta\right)
\end{equation}
%
\paragraph{E-step}
The aim of this step is to estimate the value of $Q\left(\mathbf{c}_{k}^{(i)}\right)$, which can be represented as $p\left(\mathbf{c}_{k}^{(i)} \mid \mathbf{x}^{(i)}, \theta\right)$. 
To this end, the Sinkhorn algorithm \cite{cuturi2013sinkhorn} is employed to generate pseudo labels in line with the prior class distribution, as opposed to maximizing the prediction with $\hat{y}^{(i)}=$ $\operatorname{argmax}_{\mathbf{c}_k \in \mathbf{C}} \mathbf{c}_{k}^{\top} \cdot f_\theta\left(\mathbf{x}^{(i)}\right)$.
In order to formalize the optimal transport problem for proper label assignments, consider the following setup. At each training step, we aim to search for pseudo-labels $\mathbf{A}$ that closely approximate the current classifier's predictions $\mathbf{P}$, while adhering to specific constraints:
\begin{equation}\label{eq:ot}
   \begin{split}
 {\min _{\mathbf{A} \in \Delta} E(\mathbf{A}, \mathbf{P})=\langle\mathbf{A},-\log (\mathbf{P})\rangle}
   \\
   \text { s.t. } \Delta=\left\{\mathbf{A}^{\top} \mathbf{1}_{m}=\mathbf{r}, \mathbf{A} \mathbf{1}_{L}=\mathbf{w}\right\} 
\end{split}
\end{equation}
where $\mathbf{r}$ is an $|\mathcal{Y}_u|$-dimensional probability simplex that indicates the prior class distribution. 
Note that, here we temporarily assume we have a decent estimation of the class priors and we will describe the means of estimation in Section \ref{subsec:prior_estimation}.
The column vector $\mathbf{w} = \frac{1}{m}\mathbf{1}_m$ indicates that our $m$ training examples are sampled uniformly. 
%
To resolve Eq.~\ref{eq:ot}, we adapt the well-known Sinkhorn-Knopp algorithm \cite{cuturi2013sinkhorn} for efficient optimization.
Formally, we define a matrix $\mathbf{K}$ such that $\mathbf{K}_{i j}=\mathbf{P}_{i j}^{\lambda}$, where $\lambda>0$ is a smoothing regularization coefficient.
$\mathbf{K}$ can be efficiently computed using Sinkhorn's fixed point iteration: $\mathbf{\alpha}, \mathbf{\beta} \leftarrow \mathbf{r} . / \mathbf{K} \mathbf{\beta}, \mathbf{w} . / \mathbf{K}^{\top} \mathbf{\alpha}$, where $. /$ denotes element-wise division.
Additionally, inspired by \cite{caron2020unsupervised,he2020momentum}, we also involve a queue acceleration trick to avoid traversing the whole training set.

\paragraph{M-step}
Based on the E-step, we are ready to maximize the lower-bound in Eq.~\ref{eq:em2} with respect to $\theta$:
\begin{equation}
\small
   \label{eq:m-step}
   \begin{aligned}
&\sum_{i=1}^{n} \sum_{c_{k} \in C} Q\left(\mathbf{c}_{k}^{(i)}\right) \log p\left(\mathbf{x}^{(i)}, \mathbf{c}_{k}^{(i)} \mid \theta\right) \\
=&\sum_{i=1}^{n} \sum_{c_{k} \in C} \underbrace{\deepblue{\mathbbm{1}\left(\mathbf{x}^{(i)} \in \mathbf{c}_{k}^{(i)}\right)}}_{\deepblue{\Large{\text{Achieved by Sinkhorn algorithm in E-step}}}} \log p\left(\mathbf{x}^{(i)}, \mathbf{c}_{k}^{(i)} \mid \theta\right)
\end{aligned}
\end{equation}
And we also have
$
p\left(\mathbf{x}^{(i)}, \mathbf{c}_{k}^{(i)} \mid \theta\right)=p\left(\mathbf{x}^{(i)} \mid \mathbf{c}_{k}^{(i)}, \theta\right) p\left(\mathbf{c}_{k}^{(i)} \mid \theta\right),
$
we derive prior probability $p\left(\mathbf{c}_{k}^{(i)} \mid \theta\right)$ from class prior estimation (Section \ref{subsec:prior_estimation}). Following \cite{li2020prototypical}, we assume an isotropic Gaussian distribution around each prototype with an  the same variance $\sigma$:
%
\begin{equation}
   \label{eq:guassian}
   \begin{aligned}
&p\left(x^{(i)} \mid c_{k}^{(i)}, \theta\right) \\
=&\exp \left(\frac{-\left(v_{i}-c_{k}^{(i)}\right)^{2}}{2 \sigma^{2}}\right) / \sum_{j=1}^{|\mathcal{Y}_u|} \exp \left(\frac{-\left(v_{i}-c_{j}\right)^{2}}{2 \sigma^{2}}\right),
\end{aligned}
\end{equation}
with $\mathbf{v}^{(i)}=f_{\theta}\left(\mathbf{x}^{(i)}\right)$ and $\mathbf{x}^{(i)} \in \mathbf{c}_{k}^{(i)}$, combining the above equations, we express maximum log-likelihood estimation as:
\begin{equation}
\small
   \label{eq:guassian}
   \begin{aligned}
\theta^{*}=&\underset{\theta}{\arg \min } \sum_{i=1}^{m}-\log \frac{ \exp \left(\mathbf{v}^{(i)} \cdot \mathbf{c}_{k}^{(i)} \right)}{\sum_{j=1}^{|\mathcal{Y}_u|} \exp \left(\mathbf{v}^{(i)} \cdot \mathbf{c}_{j} \right)} - \log p\left(\mathbf{c}_{k}^{(i)} \mid \theta\right)
\end{aligned}
\end{equation}
The prototype loss of the unlabeled data is:
\begin{equation}
\small
   \label{eq:proto_loss}
   \begin{aligned}
\mathcal{L}_{proto}^{(i)}=-\log \frac{ \exp \left(\mathbf{v}^{(i)} \cdot \mathbf{c}_{k}^{(i)} \right)}{\sum_{j=1}^{|\mathcal{Y}_u|} \exp \left(\mathbf{v}^{(i)} \cdot \mathbf{c}_{j}\right)} - \log p\left(\mathbf{c}_{k}^{(i)} \mid \theta\right)
\end{aligned}
\end{equation}


\subsection{Representation Improvement}
\label{subsec:label_part}
To enhance model representation, we adopt unsupervised contrastive learning \cite{chen2020simple, he2020momentum} for unlabeled data and supervised contrastive learning \cite{khosla2020supervised} for labeled data, as in \cite{sunopencon,cao2022open}. Specifically, let $\mathbf{v}_{i}$ and $\mathbf{v}_{i}^{\prime}$ represent features from two views (random augmentations) of the same image within a mini-batch $B$. The instance-level unsupervised contrastive loss and the supervised constrastive loss are defined as follows:
\begin{equation}
\small
   \begin{aligned}
      \mathcal{L}^{(i)}_{ins}=\frac{1}{|B_u|}\sum_{i \in B_u}-\log \frac{\exp \left(\mathbf{v}_{i} \cdot \mathbf{v}_{i}^{\prime} / \tau\right)}{\sum_{i}^{i \neq j} \exp \left(\mathbf{v}_{i} \cdot \mathbf{v}_{j} / \tau\right)},
   \end{aligned}
\end{equation}
\begin{equation}
   \small
   \mathcal{L}^{(i)}_{sup}=\frac{1}{\left|B_{l}\right|}\sum_{i \in B_{l}}\frac{1}{|\mathcal{N}^{(i)|}} \sum_{q \in \mathcal{N}^{(i)}} -\log \frac{\exp \left(\mathbf{v}_{i} \cdot \mathbf{v}_{q} / \tau\right)}{\sum_{i}^{i \neq j}\exp \left(\mathbf{v}_{i} \cdot \mathbf{v}_{j} / \tau\right)},
   \end{equation}
%
%
where $B_u$ and $B_l$ represent the unlabeled and labeled subsets of mini-batch $B$, respectively. Furthermore, $\mathcal{N}^{(i)}$ refers to the indices of other images in the batch sharing the same label, and $\tau$ signifies a temperature parameter.

\subsection{Overall Loss Objective}
\label{subsec:overall_loss}
The overall loss objective is a weighted sum of the unsupervised and supervised contrastive losses:
\begin{equation}
   \label{eq:overall}
   \small
\mathcal{L_\text{overall}}= \underbrace{\mathcal{L}_{ins}+\lambda_{proto}\mathcal{L}_{proto}}_{\mathcal{L}_{unlabeled}} + \underbrace{\lambda_{sup}\mathcal{L}_{sup}}_{\mathcal{L}_{labeled}},
\end{equation}
where $\lambda_{proto}$ and $\lambda_{p}$ are a hyper-parameter controlling the relative weight of the supervised loss. 
The contrastive loss derived from the unlabeled data can be examined within two distinct hierarchical tiers: the instance level, referred to as instance contrastive loss, $\mathcal{L}_{ins}$, and the category level, identified as prototype contrastive loss, $\mathcal{L}_{proto}$. Conversely, for the labeled data, only the category level is taken into account, which is referred to as supervised contrastive loss, ${L}_{sup}$.




%
\subsection{Class Prior Estimation \& Prototype Updates}
\label{subsec:prior_estimation}
%
\paragraph{Moving-average distribution update} 
We suggest employing model predictions for class prior estimation following \cite{guo2017calibration}. However, due to potential inaccuracies and biases in early training stages, we propose a moving-average update mechanism to enhance reliability.
Starting with a uniform class prior $\mathbf{r}=[1 / C, \ldots, 1 / C]$, we iteratively refine the distribution per epoch.
   \begin{gather*}
    \small
      \mathbf{r} := \mu \mathbf{r}+(1-\mu) \mathbf{z}, \\
      \quad \text { where } \mathbf{z}_{j}=\frac{1}{n} \sum_{i=1}^{n} \mathbbm{1}\left(j=\arg \max _{j^{\prime}} f_{j^{\prime}}\left(\mathbf{x}_{i}\right)\right), \mu \in[0,1]
   \end{gather*}
The class prior is continuously updated through a linear function, resulting in more stable training dynamics. As the training progresses, the model's accuracy improves, making the estimated distribution increasingly dependable.
\paragraph{Momentum prototypes update}
A canonical approach for updating prototype embeddings is computationally expensive. To reduce training latency, we employ a moving-average strategy \cite{li2020mopro} for updating class-conditional prototype vectors:
$$
\small
\mathbf{c}_{k} := \mu \mathbf{c}_{k}+(1-\mu) \mathbf{v}_k,
$$
where the prototype $\mathbf{c}_{k}$ of class $k$-th is defined by the moving average of the normalized embeddings $\mathbf{v_k}$, whose predicted class conforms to $k$. 
For both iterate updates, we employ the same hyperparameter $\mu$ and we upate prototypes and class prior each epoch.

\begin{table*}[]
   \label{tab:cifar100}
   \centering
   \setlength{\tabcolsep}{12pt}
   \caption{Performance comparison of various methods on CIFAR100 under different imbalanced factors $\rho$ ($\rho=0.5$ is original GCD setting). 
 Our method consistently outperforms others on novel class, showcasing its effectiveness in handling class imbalance.
We report the mean and standard deviation of the
clustering accuracy across 3 runs for multiple methods. 
The higher mean
  value is presented in bold, while the results within standard deviation of the average accuracy are not bolded.}
   \label{tab:CIFAR100}
   \begin{tabular}{@{}lccccl@{}}
      \toprule
   \multirow{2}{*}{IMF} & \multirow{2}{*}{Metrics} & \multicolumn{4}{c}{Methods}                     \\ \cmidrule(l){3-6} 
                                       &                          & GCD                  & \multicolumn{1}{c}{ORCA} & \multicolumn{1}{c}{OpenCon} & \multicolumn{1}{c}{Ours} \\
                                       \midrule
   \multirow{5}{*}{$\rho=0.5$}
   & All                   & 45.41\tiny{$\pm$0.13} & 55.68\tiny{$\pm$0.33}& 51.85\tiny{$\pm$0.63} & 53.51\tiny{$\pm$0.26} \\
   & Known                  & 67.61\tiny{$\pm$0.12} & 66.41\tiny{$\pm$0.31} & 69.07\tiny{$\pm$0.29} & 68.09\tiny{$\pm$0.13} \\
   & Unknown-aware              & 34.31\tiny{$\pm$0.22} & 42.63\tiny{$\pm$0.67} & 45.76\tiny{$\pm$0.32} & \textbf{47.92}\tiny{$\pm$0.33} \blue{\small{($+$ 2.16)}} \\
   & Unknown-agnostic              & 18.12\tiny{$\pm$0.34} & 38.95\tiny{$\pm$0.76} & 42.11\tiny{$\pm$0.54} & \textbf{46.22}\tiny{$\pm$0.33} \blue{\small{($+$ 4.11)}} \\
                               \midrule
   \multirow{5}{*}{$\rho=1$}   
   & All                   & 48.36\tiny{$\pm$0.08} & 47.37\tiny{$\pm$0.52} & 53.20\tiny{$\pm$0.33} & 54.06\tiny{$\pm$0.45} \\					
   &	Known                 & 71.48\tiny{$\pm$0.24} & 65.14\tiny{$\pm$0.17} & 68.00\tiny{$\pm$0.06} & 67.98\tiny{$\pm$0.37} \\					
   &	Unknown-aware          & 25.24\tiny{$\pm$0.07} & 34.93\tiny{$\pm$1.04} & 43.58\tiny{$\pm$0.34} & 43.39\tiny{$\pm$0.59} \\					
   &	Unknown-agnostic             & 12.02\tiny{$\pm$0.46} & 29.61\tiny{$\pm$1.02} & 38.40\tiny{$\pm$0.65} & \textbf{40.72}\tiny{$\pm$0.9} \blue{\small{($+$ 2.32)}}\\			
                               \midrule
   \multirow{5}{*}{$\rho=5$}  
   & All      & 63.13\tiny{$\pm$0.12} & 56.37\tiny{$\pm$0.12} & 61.84\tiny{$\pm$0.29} & 59.48\tiny{$\pm$0.48}\\ 
   & Known     & 70.96\tiny{$\pm$0.13} & 64.40\tiny{$\pm$0.11} & 69.37\tiny{$\pm$0.23} & 67.82\tiny{$\pm$0.06}\\
   & Unknown-aware & 24.04\tiny{$\pm$0.12} & 25.36\tiny{$\pm$0.40} & 35.06\tiny{$\pm$0.34} & \textbf{37.87}\tiny{$\pm$1.59} \blue{\small{($+$ 2.21)}}\\
   & Unknown-agnostic & 7.18\tiny{$\pm$0.34}  & 16.18\tiny{$\pm$0.23} & 24.21\tiny{$\pm$0.67} & \textbf{27.64}\tiny{$\pm$2.26} \blue{\small{($+$ 3.43)}}\\
   \midrule
   \multirow{5}{*}{$\rho=10$}  
   & All      & 66.21\tiny{$\pm$0.23} & 59.61\tiny{$\pm$0.27} & 65.05\tiny{$\pm$0.29} & 63.21\tiny{$\pm$0.05}\\ 
   & Known     & 70.36\tiny{$\pm$0.31} & 64.27\tiny{$\pm$0.32} & 69.80\tiny{$\pm$0.23} & 67.82\tiny{$\pm$0.07}\\
   & Unknown-aware & 24.74\tiny{$\pm$0.45} & 26.21\tiny{$\pm$0.69} & 33.01\tiny{$\pm$0.34} & \textbf{34.87}\tiny{$\pm$0.70} \blue{\small{($+$ 1.76)}}\\
   & Unknown-agnostic & 7.28\tiny{$\pm$0.21}  & 13.01\tiny{$\pm$0.34} & 17.57\tiny{$\pm$0.67} & \textbf{21.68}\tiny{$\pm$0.29} \blue{\small{($+$ 4.12)}}\\
                               \bottomrule                            
   \end{tabular}
   \end{table*}

\begin{table*}[]
   \centering
   \setlength{\tabcolsep}{12pt}
   \caption{Performance comparison of various methods on ImageNet100 under different imbalanced factors $\rho$, ($\rho=0.5$ is original GCD setting). 
   Our method consistently outperforms others on novel class under imbalanced settings and achieves competitive performance on the balanced setting.
  We report the mean and standard deviation of the
  clustering accuracy across 3 runs for multiple methods. 
  The higher mean
  value is presented in bold, while the results within standard deviation of the average accuracy are not bolded.}
   \label{tab:ImageNet100}
   \begin{tabular}{@{}lcclll@{}}
      \toprule
   \multirow{2}{*}{IMF} & \multirow{2}{*}{Metrics} & \multicolumn{4}{c}{Methods}                     \\ \cmidrule(l){3-6} 
                                       &                          & GCD                  & \multicolumn{1}{c}{ORCA} & \multicolumn{1}{c}{OpenCon} & \multicolumn{1}{c}{Ours} \\
                                       \midrule
   \multirow{5}{*}{$\rho=0.5$} 
   & All                   & 77.12\tiny{$\pm$0.56} & 74.93\tiny{$\pm$0.34} & 82.22\tiny{$\pm$0.56} & 81.90\tiny{$\pm$0.67}\\ 
   & Known                  & 87.02\tiny{$\pm$0.36} & 89.21\tiny{$\pm$0.06} & 90.65\tiny{$\pm$0.04} & \textbf{91.19}\tiny{$\pm$0.15}\\
   & Unknown-aware              & 57.48\tiny{$\pm$0.76} & 67.18\tiny{$\pm$0.27} & 78.12\tiny{$\pm$0.80} & 77.89\tiny{$\pm$0.88}\\
   & Unknown-agnostic              & 42.68\tiny{$\pm$0.77} & 65.43\tiny{$\pm$0.37} & 78.01\tiny{$\pm$0.83} & 77.79\tiny{$\pm$0.97}\\
                                 \midrule
   \multirow{5}{*}{$\rho=1$}   
   & All   & 63.06\tiny{$\pm$0.66} & 68.01\tiny{$\pm$0.31}  & 82.44\tiny{$\pm$0.24} &  82.34\tiny{$\pm$0.39} \\
   & Known & 88.80\tiny{$\pm$0.43} & 88.99\tiny{$\pm$0.05}  & 90.62\tiny{$\pm$0.12} &  90.56\tiny{$\pm$0.18} \\
   & Unknown-aware  & 37.29\tiny{$\pm$1.11} & 47.68\tiny{$\pm$0.70}  & 74.45\tiny{$\pm$0.36} &  74.56\tiny{$\pm$0.79} \\
   & Unknown-agnostic  & 31.16\tiny{$\pm$1.48}  & 47.28\tiny{$\pm$0.66}  & 74.35\tiny{$\pm$0.38} &   74.24\tiny{$\pm$0.89}\\
                                 \midrule
           \multirow{5}{*}{$\rho=5$}   
   & All      & 77.29\tiny{$\pm$0.17}  & 76.79\tiny{$\pm$0.13} & 81.87\tiny{$\pm$0.23} & \textbf{83.01}\tiny{$\pm$0.11}\\
   & Known     & 87.51\tiny{$\pm$0.21}  & 89.02\tiny{$\pm$0.12} &  \textbf{90.77}\tiny{$\pm$0.10} & 88.89\tiny{$\pm$0.06}\\
   & Unknown-aware & 23.83\tiny{$\pm$0.24}  & 20.19\tiny{$\pm$0.55} & 39.76\tiny{$\pm$1.4} & \textbf{54.35}\tiny{$\pm$0.38} \blue{\small{($+$ 14.59)}}\\
   & Unknown-agnostic & 14.80\tiny{$\pm$0.23}  & 16.42\tiny{$\pm$0.73} & 37.93\tiny{$\pm$1.3} & \textbf{53.97}\tiny{$\pm$0.38} \blue{\small{($+$ 16.04)}}\\
   \midrule
   \multirow{5}{*}{$\rho=10$}   
   & All      & 83.32\tiny{$\pm$0.15}  & 81.98\tiny{$\pm$0.03} & \textbf{84.60}\tiny{$\pm$0.10} & 83.23\tiny{$\pm$0.17}\\
   & Known     & 89.71\tiny{$\pm$0.03}  & 89.24\tiny{$\pm$0.02} & \textbf{90.85}\tiny{$\pm$0.12} & 87.48\tiny{$\pm$0.23}\\
   & Unknown-aware & 22.17\tiny{$\pm$0.18}  & 16.94\tiny{$\pm$0.07} & 26.70\tiny{$\pm$0.11} & \textbf{42.62}\tiny{$\pm$0.50} \blue{\small{($+$ 15.92)}}\\
   & Unknown-agnostic & 11.47\tiny{$\pm$0.28}  & 10.31\tiny{$\pm$0.22} & 22.89\tiny{$\pm$0.27} & \textbf{41.23}\tiny{$\pm$0.58} \blue{\small{($+$ 18.34)}}\\   
   \bottomrule               
   \end{tabular}
   \end{table*}

\section{Experiments}
In this section, we conduct an experimental analysis of the proposed ImbaGCD approach under both original and varying imbalanced scenarios.
\subsection{Setup}
\paragraph{Datasets} 
We conducted experiments on three datasets, including CIFAR10, CIFAR100 and ImageNet-100 (where ImageNet-100 is a subsampled version of the ImageNet dataset with 100 classes), the classes were divided into 50\% known and 50\% unknown classes.
And we randomly selected 50\% of the known class samples as the labeled dataset, which is balanced. 
However, the sample sizes of known and unknown classes in the unlabeled set are imbalanced, which is adjusted based on the imbalanced factor $\rho$. 
Here, $\rho$ was defined as the ratio of the sample sizes of the known and unknown classes, i.e., $\rho=\frac{n_{k}}{n_{u}}$, where $n_{k}$ and $n_{u}$ represent the sample sizes of known and unknown classes, respectively, and $\rho \in\{0.5, 1, 5, 10\}.$ 
It is worth noting that previous works often selected 50\% of the known class samples as the labeled dataset, and the remaining samples were used for the unlabeled set, resulting in a fixed imbalance ratio of $\rho=0.5$.

\paragraph{Evaluation metrics} 
We adopt the evaluation strategy outlined in \cite{cao2022open,sunopencon,vaze2022generalized} and report the following metrics: (1) overall accuracy across all classes, (2) classification accuracy for known classes. Additionally, we evaluate novel data using two distinct evaluation settings: (3) \textbf{unknown-aware} and (4) \textbf{unknown-agnostic}. As described in \cite{sunopencon}, the accuracy for novel classes and all classes is determined by solving an optimal assignment problem using the Hungarian algorithm \cite{kuhn1955hungarian}.
In the unknown-aware evaluation, we adhere to the established GCD methods by first isolating all unlabeled samples associated with unknown classes. We then perform clustering specifically within these unknown class categories. 
However, this approach may not accurately reflect real-world scenarios, as directly distinguishing between known and unknown classes within unlabeled data is often impractical.
To address this limitation, we also report unknown-agnostic accuracy, which refrains from using any information to differentiate between known and unknown classes within unlabeled data, providing a more realistic and unbiased evaluation metric.

\paragraph{Implementation details}
We utilize ResNet-18 as the backbone architecture for CIFAR-100 and ResNet-50 for ImageNet-100. In addition, we introduce a trainable two-layer MLP projection head that maps the features from the penultimate layer to a lower-dimensional space $\mathbb{R}^{d}(d=128)$. This projection technique has proven effective for contrastive loss \cite{chen2020simple}. In line with the methods proposed in \cite{cao2022open,sunopencon}, we implement regularization by calculating the KL-divergence between the predicted label distribution and the class prior, which helps to improve model stability.

For both CIFAR10/100 and ImageNet-100, the model undergoes training for 80 and 120 epochs, respectively, with a batch size of 512. We employ stochastic gradient descent with a momentum of 0.9 and a weight decay of $10e-4$. The learning rate commences at 0.02 and undergoes decay by a factor of 10 at the 50\% and 75\% stages of the training process. The momentum for updating the prototype and class prior, represented as $\mu$, is consistently maintained at 0.99.

\begin{figure}[]
   \vspace{-0.0cm}
   \centering
   \includegraphics[width=0.7\linewidth]{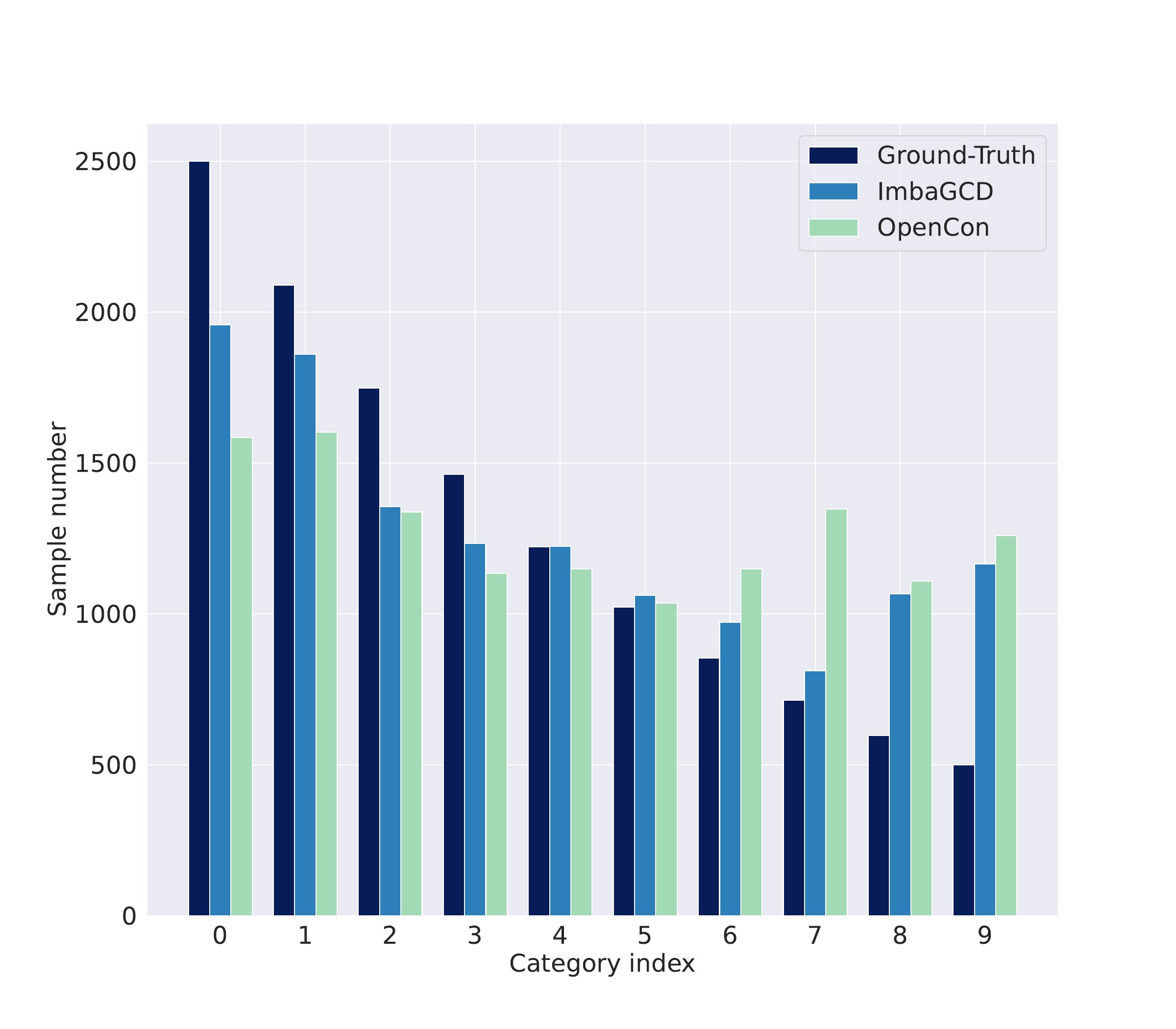}
   \caption{Comparison of the predicted sample numbers for each class on the CIFAR-10 dataset, using an exponential decreasing strategy with an imbalance factor ($\rho$) of 5. }
   \label{fig:cifar10_exp5}
\end{figure}

\begin{figure}[]
   \vspace{-0.0cm}
   \centering
   \includegraphics[width=0.7\linewidth]{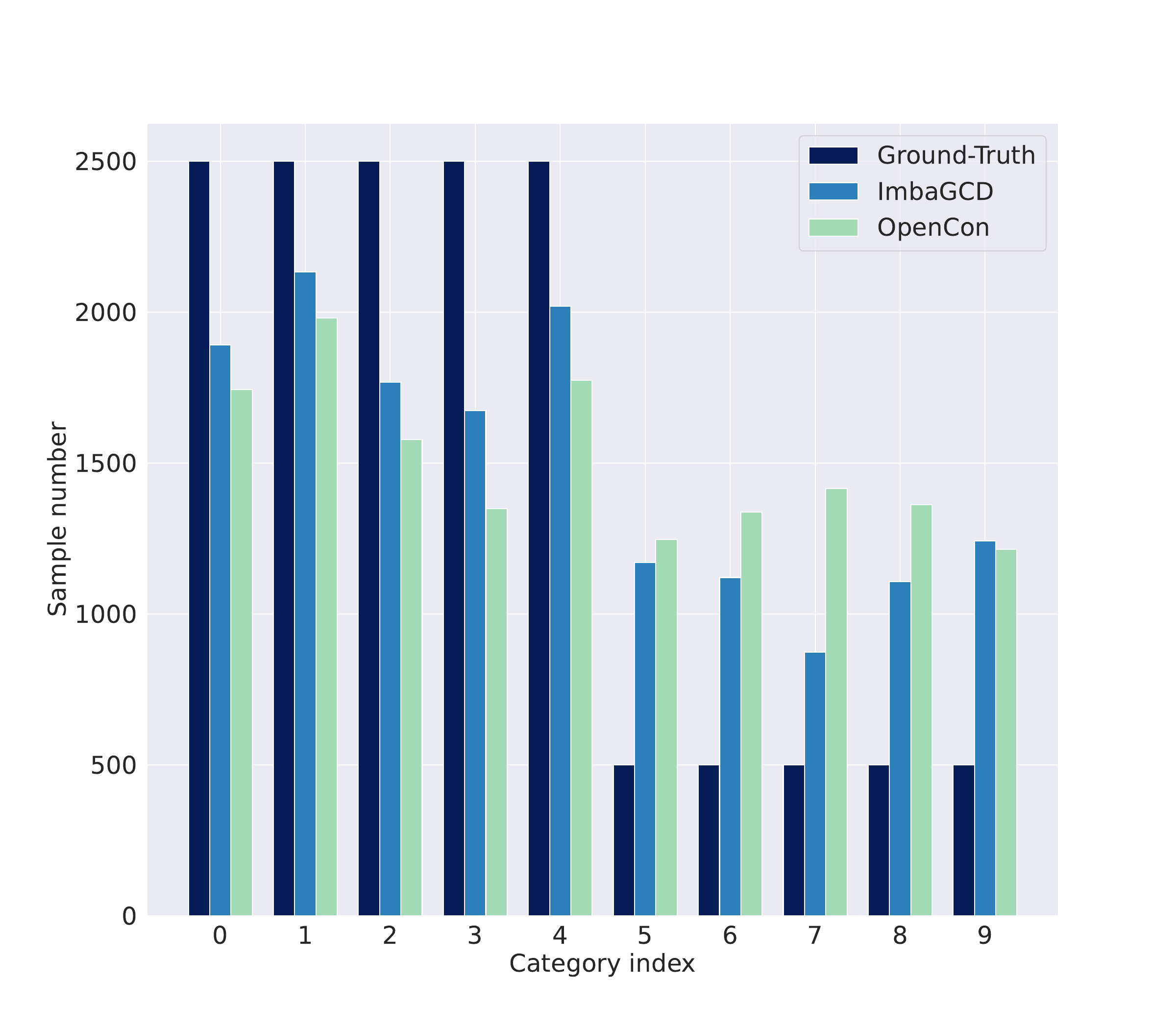}
   \caption{Comparison of the predicted sample numbers for each class on the CIFAR-10 dataset, using an step decreasing strategy with an imbalance factor ($\rho$) of 5. }
   \label{fig:cifar10_step5}
\end{figure}

\subsection{Compared with SOTA}
\paragraph{Baselines} 
In our study, we evaluate our approach against three state-of-the-art GCD methods serving as baselines:
(1) GCD \cite{vaze2022generalized}, the pioneering framework for GCD, efficiently identifies and clusters unknown categories in high-dimensional data without exhaustive labeling;
(2) ORCA \cite{cao2022open}, combines supervised and unsupervised learning to effectively utilize labeled and unlabeled data, addressing class imbalance in open-world settings;
(3) OpenCon \cite{sunopencon}, employs contrastive learning to maximize similarity between positive pairs while minimizing it for negative pairs, enhancing adaptability to new categories with minimal supervision.
We evaluate our method against the aforementioned baselines in various scenarios: original GCD setting, balanced setting, and multiple imbalanced settings, allowing for a comprehensive comparison across diverse conditions.



\paragraph{ImbaGCD achieves SOTA performance}
In Tables \ref{tab:CIFAR100} and \ref{tab:ImageNet100}, ImbaGCD demonstrates a significant performance advantage over its competitors on both CIFAR-100 and ImageNet datasets, particularly in the novel classes. Specifically, on the CIFAR-100 dataset, our method achieves an improvement of approximately 2 - 4\% over the best baseline in terms of both unknown-aware and unknown-agnostic accuracy across a range of imbalanced settings.
Regarding the ImageNet-100 dataset, ImbaGCD surpasses the baseline performance by approximately 14-16\% and 15-19\% under $\rho=5$ and $\rho=10$ settings, respectively. Additionally, it attains competitive results in balanced and original GCD settings ($\rho=0.5$).
It is important to highlight that the unknown-agnostic evaluation presents a greater challenge compared to the unknown-aware evaluation, particularly in the context of highly imbalanced settings. Our method exhibits more substantial improvements under these challenging evaluation conditions, with the enhancements in unknown-agnostic performance being more prominent than those in unknown-aware performance.

\begin{table}[]
    \centering
    \centering
    \setlength{\tabcolsep}{5pt}
 \caption{Ablation study on loss componet in CIFAR100. Known class, unknown-aware(Un1), and unknown-agnostic (Un2) accuracies.}
    \label{tab:ablation}
    \begin{tabular}{@{}lllll@{}}
       \toprule
    \multicolumn{1}{l}{IMF}        & \multicolumn{1}{l}{Loss} & \multicolumn{1}{l}{Known} & \multicolumn{1}{l}{Un1} & \multicolumn{1}{l}{Un2} \\
    \midrule
    \multirow{4}{*}{$\rho=0.5$} 
    & w/o $L_{sup}$          &  42.86 & 46.51 & 46.91 \\
    & w/o $L_{ins}$          &  64.96 & 17.40 & 12.09\\
    & w/o $L_{proto}$        &  68.36 & 46.15 &	44.21\\
    & Ours  & 68.09\tiny{$\pm$0.13} & {47.92}\tiny{$\pm$0.33} & {46.22}\tiny{$\pm$0.33} \\
    \midrule
    \multirow{4}{*}{$\rho=1$}   
    & w/o $L_{sup}$          &   43.20 & 42.90 & 39.96\\
    & w/o $L_{ins}$          &  63.68 & 17.54 & 6.43\\
    & w/o $L_{proto}$        &  67.57 & 42.91 &	38.27 \\
    & Ours & 67.98\tiny{$\pm$0.37} & {43.39}\tiny{$\pm$0.59} & {38.76}\tiny{$\pm$0.90} \\
    \midrule
    \multirow{4}{*}{$\rho=5$}  
    & w/o $L_{sup}$          &  31.01 & 30.40 & 9.25 \\
    & w/o $L_{ins}$          &  52.23 & 18.24 & 12.07\\
    & w/o $L_{proto}$        &  66.80 &	32.72 &	21.20  \\
    & Ours & 67.82\tiny{$\pm$0.06} & \textbf{37.87}\tiny{$\pm$1.59} & \textbf{27.64}\tiny{$\pm$2.26}\\
    \midrule
    \multirow{4}{*}{$\rho=10$} 
    & w/o $L_{sup}$          &   30.33 & 29.92 & 4.50\\
    & w/o $L_{ins}$          &  51.75 & 20.88 & 13.42\\
    & w/o $L_{proto}$        &  66.52 &	32.56	 & 14.76 \\
    & Ours & 67.82\tiny{$\pm$0.07} & \textbf{34.87}\tiny{$\pm$0.70} & \textbf{21.68}\tiny{$\pm$0.29}\\
    \bottomrule                            
    \end{tabular}
    \label{tab:ablation}
\end{table}

\subsection{Ablation Study}
\paragraph{Class distribution prediction}
ImbaGCD outperforms the state-of-the-art OpenCon in predicting class distributions for different decreasing strategies on the CIFAR10.
We conduct experiments on two decreasing types: exponential-decreasing (Figure \ref{fig:cifar10_exp5}) and step-decreasing (Figure \ref{fig:cifar10_step5}). The $x$-axis represents the class index, while the $y$-axis denotes the sample number of each class.
Upon comparing our results with those of OpenCon \cite{sunopencon} and the ground-truth sample numbers, it becomes evident that our method consistently attains a superior class distribution. Furthermore, the deviations between the predictions made by our method and the ground-truth values are consistently smaller than those observed for OpenCon across all classes. These findings underscore the effectiveness and robustness of our approach in predicting class distributions.


\paragraph{Analysis of the loss components}
Recall our objective function in Eq. \ref{eq:overall} has three components. We perform an ablation study (Table \ref{tab:ablation}, with Un1 as Un1accuracy and Un2 as Un2 accuracy) to analyze their contributions. The ImbaGCD model is modified by removing: $L_{sup}$, ${L}_{ins}$, and $L_{proto}$. This study aims to understand each component's impact on performance.
We make several observations from our ablation study:
\begin{itemize}
   \vspace{-2.5mm}
  \item When there are more unknown class samples (e.g., $\rho=0.5, 1$), removing $L_{sup}$ mainly affects known class accuracy with less impact on Un1 and Un2 ($<1\%$). For fewer unknown samples (e.g., $\rho=5, 10$), reductions in known class accuracy cause significant drops in Un1 ($\sim5-7\%$) and Un2 ($17-18\%$).
   \vspace{-2.5mm}
   \item For $L_{ins}$, at $\rho=0.5$ and $\rho=1$, the known class accuracy decreases slightly, while the declines in Un1 and Un2 are more pronounced ($\sim26-32\%$). However, when $\rho=5$ and $\rho=10$, there is a noticeable drop in all metrics (known, Un1, and Un2)
   \vspace{-2.5mm}
   \item For $L_{proto}$, the effects are not significant when $\rho=0.5$ and $\rho=1$. However, there is a notable improvement in Un1 ($\sim2-5\%$) and Un2 ($\sim6-7\%$) when the $L_{proto}$ component is included.
\end{itemize}

\section{Related Work}
\paragraph{Generalized category discovery (GCD)} This problem extends NCD \cite{han2019learning, han2021autonovel, fini2021unified, li2022closer} by considering unlabeled data from both known and novel classes \cite{vaze2022generalized}. 
GCD addresses this challenge through semi-supervised contrastive learning on large-scale pre-trained visual transformers (ViT) followed by constraint KMeans \cite{arthur2007k}. 
Concurrently, ORCA \cite{cao2022open} proposes an uncertainty adaptive margin loss to reduce intra-class variances between known and novel classes.
Opencon \cite{sunopencon} proposes a contrastive learning frameworks which selects the positive and negative pair via a moving average prototype.
Despite the prevalence of class imbalance, many existing works in this area overlook its impact. Our work contributes to the field by addressing this gap.
%

\paragraph{Learning with class-imbalanced data} 
Real-world datasets often exhibit a long-tailed label distribution \cite{mahajan2018exploring, van2018inaturalist}, complicating standard DNN training and generalization \cite{dong2018imbalanced, ren2018learning, wang2017learning}. To address class imbalance, approaches include (a) re-weighting loss functions class-wise \cite{cao2019learning, park2021influence,li2021autobalance}, and (b) re-sampling datasets for balanced training distribution \cite{chawla2002smote, byrd2019effect}. Both methods perform better when applied in later training stages for DNNs \cite{kang2019decoupling, tang2020long}. However, they assume full supervision, prompting studies on weak-supervision, such as semi-supervised learning \cite{kim2020distribution, wei2021crest}, where \cite{kim2020distribution} requires ground-truth class priors and \cite{wei2021crest} estimates them from labeled data. 
Our work, however, presents a greater challenge as the imbalances occur between known and unknown classes, with no prior class distribution information available to estimate the known class prior distribution.


\section{Conclusion}
In this paper, we present the significant and challenging problem of Imbalanced Generalized Category Discovery (ImbaGCD), characterized by an imbalanced distribution of unlabeled data. To tackle this issue, we develop ImbaGCD, a novel and robust optimal transport-based expectation maximization framework.
Our extensive experimental evaluation encompasses varying settings, including balanced and multiple imbalanced scenarios. The results demonstrate that our proposed method consistently outperforms state-of-the-art approaches across diverse imbalanced settings. Furthermore, ImbaGCD exhibits competitive performance in both balanced and original GCD settings, highlighting its adaptability and effectiveness across a range of situations.
These findings establish ImbaGCD as a highly capable and versatile solution for addressing the ImbaGCD problem, paving the way for further advancements in GCD.

{\small
\bibliographystyle{ieee_fullname}
\bibliography{ncd}

\begin{thebibliography}{10}\itemsep=-1pt

\bibitem{arthur2007k}
David Arthur and Sergei Vassilvitskii.
\newblock K-means++ the advantages of careful seeding.
\newblock In {\em Proceedings of the eighteenth annual ACM-SIAM symposium on
  Discrete algorithms}, pages 1027--1035, 2007.

\bibitem{buda2018systematic}
Mateusz Buda, Atsuto Maki, and Maciej~A Mazurowski.
\newblock A systematic study of the class imbalance problem in convolutional
  neural networks.
\newblock {\em Neural Networks}, 106:249--259, 2018.

\bibitem{byrd2019effect}
Jonathon Byrd and Zachary Lipton.
\newblock What is the effect of importance weighting in deep learning?
\newblock In {\em International conference on machine learning}, pages
  872--881. PMLR, 2019.

\bibitem{cao2022open}
Kaidi Cao, Maria Brbic, and Jure Leskovec.
\newblock Open-world semi-supervised learning.
\newblock In {\em ICLR}, 2020.

\bibitem{cao2019learning}
Kaihua Cao, Colin Wei, Adrien Gaidon, Nikos Arechiga, and Tengyu Ma.
\newblock Learning imbalanced datasets with label-distribution-aware margin
  loss.
\newblock {\em Advances in Neural Information Processing Systems},
  32:1565--1576, 2019.

\bibitem{caron2020unsupervised}
Mathilde Caron, Ishan Misra, Julien Mairal, Priya Goyal, Piotr Bojanowski, and
  Armand Joulin.
\newblock Unsupervised learning of visual features by contrasting cluster
  assignments.
\newblock {\em Advances in Neural Information Processing Systems},
  33:9912--9924, 2020.

\bibitem{chapelle2009semi}
Olivier Chapelle, Bernhard Scholkopf, and Alexander Zien.
\newblock Semi-supervised learning (chapelle, o. et al., eds.; 2006)[book
  reviews].
\newblock {\em IEEE Transactions on Neural Networks}, 20(3):542--542, 2009.

\bibitem{chawla2002smote}
Nitesh~V Chawla, Kevin~W Bowyer, Lawrence~O Hall, and W~Philip Kegelmeyer.
\newblock Smote: Synthetic minority over-sampling technique.
\newblock {\em Journal of Artificial Intelligence Research}, 16:321--357, 2002.

\bibitem{chen2020simple}
Ting Chen, Simon Kornblith, Mohammad Norouzi, and Geoffrey Hinton.
\newblock A simple framework for contrastive learning of visual
  representations.
\newblock In {\em International Conference on Machine Learning}, pages
  1597--1607. PMLR, 2020.

\bibitem{chi2021meta}
Haoang Chi, Feng Liu, Wenjing Yang, Long Lan, Tongliang Liu, Bo Han, Gang Niu,
  Mingyuan Zhou, and Masashi Sugiyama.
\newblock Meta discovery: Learning to discover novel classes given very limited
  data.
\newblock In {\em ICLR}, 2021.

\bibitem{cui2019class}
Yin Cui, Menglin Jia, Tsung-Yi Lin, Yang Song, and Serge Belongie.
\newblock Class-balanced loss based on effective number of samples.
\newblock In {\em Proceedings of the IEEE/CVF Conference on Computer Vision and
  Pattern Recognition}, pages 9268--9277, 2019.

\bibitem{cuturi2013sinkhorn}
Marco Cuturi.
\newblock Sinkhorn distances: lightspeed computation of optimal transport.
\newblock In {\em Proceedings of the 26th International Conference on Neural
  Information Processing Systems-Volume 2}, pages 2292--2300, 2013.

\bibitem{dong2018imbalanced}
Qi Dong, Shaogang Gong, and Xiatian Zhu.
\newblock Imbalanced deep learning by minority class incremental rectification.
\newblock {\em IEEE transactions on pattern analysis and machine intelligence},
  41(6):1367--1381, 2018.

\bibitem{fini2021unified}
Enrico Fini, Enver Sangineto, St{\'e}phane Lathuili{\`e}re, Zhun Zhong, Moin
  Nabi, and Elisa Ricci.
\newblock A unified objective for novel class discovery.
\newblock In {\em ICCV}, 2021.

\bibitem{guo2017calibration}
Chuan Guo, Geoff Pleiss, Yu Sun, and Kilian~Q Weinberger.
\newblock On calibration of modern neural networks.
\newblock In {\em Proceedings of the 34th International Conference on Machine
  Learning (ICML)}, 2017.

\bibitem{hady2013semi}
Mohamed Farouk~Abdel Hady and Friedhelm Schwenker.
\newblock Semi-supervised learning.
\newblock {\em Handbook on Neural Information Processing}, pages 215--239,
  2013.

\bibitem{han2021autonovel}
Kai Han, Sylvestre-Alvise Rebuffi, Sebastien Ehrhardt, Andrea Vedaldi, and
  Andrew Zisserman.
\newblock Autonovel: Automatically discovering and learning novel visual
  categories.
\newblock {\em IEEE Transactions on Pattern Analysis and Machine Intelligence},
  2021.

\bibitem{han2019learning}
Kai Han, Andrea Vedaldi, and Andrew Zisserman.
\newblock Learning to discover novel visual categories via deep transfer
  clustering.
\newblock In {\em CVPR}, 2019.

\bibitem{he2009learning}
Haibo He and Edwardo~A Garcia.
\newblock Learning from imbalanced data.
\newblock {\em IEEE Transactions on Knowledge and Data Engineering},
  21(9):1263--1284, 2009.

\bibitem{he2020momentum}
Kaiming He, Haoqi Fan, Yuxin Wu, Saining Xie, and Ross Girshick.
\newblock Momentum contrast for unsupervised visual representation learning.
\newblock In {\em Proceedings of the IEEE/CVF conference on computer vision and
  pattern recognition}, pages 9729--9738, 2020.

\bibitem{huang2016learning}
Chen Huang, Yining Li, Chen Change~Loy, and Xiaoou Tang.
\newblock Learning deep representation for imbalanced classification.
\newblock In {\em Proceedings of the IEEE conference on computer vision and
  pattern recognition}, pages 5375--5384, 2016.

\bibitem{kang2020decoupling}
Bingyi Kang, Saining Li, and Dacheng Tao.
\newblock Decoupling representation and classifier for long-tailed recognition.
\newblock In {\em Eighth International Conference on Learning Representations
  (ICLR)}, 2020.

\bibitem{kang2019decoupling}
Bingyi Kang, Saining Xie, Marcus Rohrbach, Zhicheng Yan, Albert Gordo, Jiashi
  Feng, and Yannis Kalantidis.
\newblock Decoupling representation and classifier for long-tailed recognition.
\newblock {\em arXiv preprint arXiv:1910.09217}, 2019.

\bibitem{khan2018cost}
Salman~H Khan, Munawar Hayat, Fatih Porikli, and Mohammed Bennamoun.
\newblock Cost-sensitive learning of deep feature representations from
  imbalanced data.
\newblock {\em IEEE transactions on neural networks and learning systems},
  29(8):3573--3587, 2018.

\bibitem{khosla2020supervised}
Prannay Khosla, Piotr Teterwak, Chen Wang, Aaron Sarna, Yonglong Tian, Phillip
  Isola, Aaron Maschinot, Ce Liu, and Dilip Krishnan.
\newblock Supervised contrastive learning.
\newblock {\em Advances in neural information processing systems},
  33:18661--18673, 2020.

\bibitem{kim2020distribution}
Jaehyung Kim, Youngbum Hur, Sejun Park, Eunho Yang, Sung~Ju Hwang, and Jinwoo
  Shin.
\newblock Distribution aligning refinery of pseudo-label for imbalanced
  semi-supervised learning.
\newblock {\em Advances in neural information processing systems},
  33:14567--14579, 2020.

\bibitem{kuhn1955hungarian}
Harold~W Kuhn.
\newblock The hungarian method for the assignment problem.
\newblock {\em Naval research logistics quarterly}, 2(1-2):83--97, 1955.

\bibitem{li2020mopro}
Junnan Li, Caiming Xiong, and Steven~CH Hoi.
\newblock Mopro: Webly supervised learning with momentum prototypes.
\newblock {\em arXiv preprint arXiv:2009.07995}, 2020.

\bibitem{li2020prototypical}
Junnan Li, Pan Zhou, Caiming Xiong, and Steven~CH Hoi.
\newblock Prototypical contrastive learning of unsupervised representations.
\newblock {\em arXiv preprint arXiv:2005.04966}, 2020.

\bibitem{li2021autobalance}
Mingchen Li, Xuechen Zhang, Christos Thrampoulidis, Jiasi Chen, and Samet
  Oymak.
\newblock Autobalance: Optimized loss functions for imbalanced data.
\newblock {\em Advances in Neural Information Processing Systems},
  34:3163--3177, 2021.

\bibitem{li2023sup}
Ziyun Li, Jona Otholt, Ben Dai, Christoph Meinel, Haojin Yang, et~al.
\newblock Supervised knowledge may hurt novel class discovery performance.
\newblock {\em Transactions of Machine Learning Research}.

\bibitem{li2022closer}
Ziyun Li, Jona Otholt, Ben Dai, Christoph Meinel, Haojin Yang, et~al.
\newblock A closer look at novel class discovery from the labeled set.
\newblock {\em arXiv preprint arXiv:2209.09120}, 2022.

\bibitem{liu2019large}
Ziwei Liu, Zhongqi Miao, Xiaohang Zhan, Jiayun Wang, Boqing Gong, and Stella~X
  Yu.
\newblock Large-scale long-tailed recognition in an open world.
\newblock In {\em Proceedings of the IEEE/CVF Conference on Computer Vision and
  Pattern Recognition}, pages 2537--2546, 2019.

\bibitem{mahajan2018exploring}
Dhruv Mahajan, Ross Girshick, Vignesh Ramanathan, Kaiming He, Manohar Paluri,
  Yixuan Li, Ashwin Bharambe, and Laurens Van Der~Maaten.
\newblock Exploring the limits of weakly supervised pretraining.
\newblock In {\em Proceedings of the European conference on computer vision
  (ECCV)}, pages 181--196, 2018.

\bibitem{park2021influence}
Seulki Park, Jongin Lim, Younghan Jeon, and Jin~Young Choi.
\newblock Influence-balanced loss for imbalanced visual classification.
\newblock In {\em Proceedings of the IEEE/CVF International Conference on
  Computer Vision}, pages 735--744, 2021.

\bibitem{ren2018learning}
Mengye Ren, Wenyuan Zeng, Bin Yang, and Raquel Urtasun.
\newblock Learning to reweight examples for robust deep learning.
\newblock In {\em International conference on machine learning}, pages
  4334--4343. PMLR, 2018.

\bibitem{rizve2022openldn}
Mamshad~Nayeem Rizve, Navid Kardan, Salman Khan, Fahad Shahbaz~Khan, and
  Mubarak Shah.
\newblock Openldn: Learning to discover novel classes for open-world
  semi-supervised learning.
\newblock In {\em Computer Vision--ECCV 2022: 17th European Conference, Tel
  Aviv, Israel, October 23--27, 2022, Proceedings, Part XXXI}, pages 382--401.
  Springer, 2022.

\bibitem{scheirer2012toward}
Walter~J Scheirer, Anderson de Rezende~Rocha, Archana Sapkota, and Terrance~E
  Boult.
\newblock Toward open set recognition.
\newblock {\em IEEE transactions on pattern analysis and machine intelligence},
  35(7):1757--1772, 2012.

\bibitem{sinkhorn1974diagonal}
Richard Sinkhorn.
\newblock Diagonal equivalence to matrices with prescribed row and column sums.
  ii.
\newblock {\em Proceedings of the American Mathematical Society}, 1974.

\bibitem{snell2017prototypical}
Jake Snell, Kevin Swersky, and Richard Zemel.
\newblock Prototypical networks for few-shot learning.
\newblock {\em Advances in neural information processing systems}, 30, 2017.

\bibitem{sunopencon}
Yiyou Sun and Yixuan Li.
\newblock Opencon: Open-world contrastive learning.
\newblock {\em Transactions of Machine Learning Research}.

\bibitem{tang2020long}
Kaihua Tang, Jianqiang Huang, and Hanwang Zhang.
\newblock Long-tailed classification by keeping the good and removing the bad
  momentum causal effect.
\newblock {\em Advances in Neural Information Processing Systems},
  33:1513--1524, 2020.

\bibitem{van2018inaturalist}
Grant Van~Horn, Oisin Mac~Aodha, Yang Song, Yin Cui, Chen Sun, Alex Shepard,
  Hartwig Adam, Pietro Perona, and Serge Belongie.
\newblock The inaturalist species classification and detection dataset.
\newblock In {\em Proceedings of the IEEE conference on computer vision and
  pattern recognition}, pages 8769--8778, 2018.

\bibitem{vaze2022generalized}
Sagar Vaze, Kai Han, Andrea Vedaldi, and Andrew Zisserman.
\newblock Generalized category discovery.
\newblock In {\em Proceedings of the IEEE/CVF Conference on Computer Vision and
  Pattern Recognition}, 2022.

\bibitem{villani2008optimal}
C{\'e}dric Villani.
\newblock {\em Optimal transport: old and new}, volume 338.
\newblock Springer Science \& Business Media, 2008.

\bibitem{villani2009optimal}
C{\'e}dric Villani et~al.
\newblock {\em Optimal transport: old and new}, volume 338.
\newblock Springer, 2009.

\bibitem{wang2017learning}
Yu-Xiong Wang, Deva Ramanan, and Martial Hebert.
\newblock Learning to model the tail.
\newblock {\em Advances in neural information processing systems}, 30, 2017.

\bibitem{wei2021crest}
Chen Wei, Kihyuk Sohn, Clayton Mellina, Alan Yuille, and Fan Yang.
\newblock Crest: A class-rebalancing self-training framework for imbalanced
  semi-supervised learning.
\newblock In {\em Proceedings of the IEEE/CVF conference on computer vision and
  pattern recognition}, pages 10857--10866, 2021.

\end{thebibliography}
}


\end{document}